\documentclass[10pt,twocolumn,letterpaper]{article}

\usepackage[pagenumbers]{cvpr} 

\usepackage{multirow}
\usepackage{amsmath}
\usepackage{arydshln}
\usepackage{mathtools, nccmath}

\DeclarePairedDelimiter{\nint}\lfloor\rceil

\usepackage[pagebackref,breaklinks,colorlinks]{hyperref}

\title{Quantization without Tears}

\author{Minghao Fu, Hao Yu, Jie Shao, Junjie Zhou, Ke Zhu and Jianxin Wu\thanks{J. Wu is the corresponding author.} \\
National Key Laboratory for Novel Software Technology, Nanjing University, China \\ School of Artificial Intelligence, Nanjing University, China \\
{\tt\small \{fumh, yuh, shaoj, zhoujj, zhuk\}@lamda.nju.edu.cn, wujx2001@gmail.com}
}

\begin{document}
\maketitle

\begin{abstract}
Deep neural networks, while achieving remarkable success across diverse tasks, demand significant resources, including computation, GPU memory, bandwidth, storage, and energy. Network quantization, as a standard compression and acceleration technique, reduces storage costs and enables potential inference acceleration by discretizing network weights and activations into a finite set of integer values. However, current quantization methods are often complex and sensitive, requiring extensive task-specific hyperparameters, where even a single misconfiguration can impair model performance, limiting generality across different models and tasks. In this paper, we propose Quantization without Tears (QwT), a method that simultaneously achieves quantization speed, accuracy, simplicity, and generality. The key insight of QwT is to incorporate a lightweight additional structure into the quantized network to mitigate information loss during quantization. This structure consists solely of a small set of linear layers, keeping the method simple and efficient. More importantly, it provides a closed-form solution, allowing us to improve accuracy effortlessly under 2 minutes. Extensive experiments across various vision, language, and multimodal tasks demonstrate that QwT is both highly effective and versatile. In fact, our approach offers a robust solution for network quantization that combines simplicity, accuracy, and adaptability, which provides new insights for the design of novel quantization paradigms. The code is publicly available at \href{https://github.com/wujx2001/QwT}{github.com/wujx2001/QwT}.

\end{abstract}

\section{Introduction}

Along with their extraordinary breakthroughs in various vision~\cite{mask_rcnn}, language~\cite{bert} and multimodal~\cite{CLIP} tasks, deep neural networks~\cite{resnet,vit} also exhibit ferocious greed for various resources: compute, GPU memory, bandwidth, storage, energy, \etc. Hence, compressing and accelerating deep nets have not only attracted interests in academia, but are also an urgent need in real-world deployments and applications.

Among various research efforts in this direction, network quantization~\cite{quantization_survey} is arguably the most practical one. Different from unstructured pruning~\cite{unstructured_pruning}, it is well supported by existing hardware. Compared to structured pruning~\cite{li2016pruning,channel_pruning}, its compression ratio is higher and its loss is relatively smaller. For example, the INT8 quantization of both FP32 weights and activations leads to roughly 4$\times$ reduction in network size and 4$\times$ speedup with almost zero accuracy loss in many applications~\cite{i-vit}, which far exceeds structured pruning. Existing methods are often categorized as Post-Training Quantization (PTQ)~\cite{CL-Calib,PD-Quant,IGQ-ViT,RepQ-ViT,PTQ4ViT,FQ-ViT} or Quantization-Aware Training (QAT)~\cite{ReactNet,ewgs,LSQ,Q-ViT,zhu2023quantized}, where the difference is whether training is required (`no' for PTQ and `yes' for QAT).

Quantization, however, is not as perfect as it seems to be. There are also obvious drawbacks and pitfalls in existing quantization methods.
\begin{itemize}
\item \textbf{The speed-accuracy dilemma}: PTQ can be thousands of times faster than QAT during the quantization process, but QAT may well be 10 percentage points higher than PTQ in accuracy during inference. 
\item \textbf{Complexity}: Quantization methods are often delicate and tricky. They often have tons of hyperparameters to tune for each specific task, and even one improperly set hyperparameter value may ruin the quantized model.
\item \textbf{Missing generality}: Relevant to their complexities, an existing method is often geared toward a specific model and/or task. Different models/tasks require different quantization methods.
\end{itemize}
Given the status quo, at this moment it does not seem unreasonable to treat the act of network quantization as an art rather than an established engineering tool.

In this paper, we propose a Quantization without Tears (QwT) method to address these drawbacks, which achieves quantization speed, accuracy, simplicity and generality \emph{simultaneously}.

The key to achieve these goals simultaneously is to slightly change the quantization paradigm. Suppose a network $M$ has the network structure $S$ and parameters $\theta$. Current quantization methods will quantize it into a model $M^{\mathbb Z}$ with the same structure $S$ $(S^{\mathbb Z}=S)$ and quantized parameters $\theta^{\mathbb Z}$ in the integer format.

Our key argument is that the quantized structure does \emph{not} need to be strictly $S$. In our QwT, it becomes $S \cup S_c$, where some extra modules $S_c$ are added to the network structure to \emph{compensate for the information loss due to quantized parameters and activations}. The extra $S_c$ thus help us achieve high accuracy.

$S_c$ does lead to extra overheads. But, it is also obvious that so long as the size and computation of $S_c$ is small or even negligible when compared to $S^{\mathbb Z}$, we achieve \emph{both speed and accuracy}. In our QwT, $S_c$ has very simple structures: only few linear layers, which renders it \emph{both simple and general}.

To be more specific, the parameters in $S_c$ can be set in \emph{closed-form with a small calibration set}, which in almost all cases leads to significantly higher accuracy than PTQ methods. 

To sum up, the contributions of this paper are:
\begin{itemize}
	\item Proposing a new paradigm for network quantization by lifting the restriction that the quantized network structure $S^{\mathbb Z}$ has to be exactly the same as that of the original network structure $S$.
	\item Proposing QwT, a simple and general quantization method without tears in this new paradigm. QwT achieves speed, accuracy, simplicity and generality simultaneously.
\end{itemize}

Extensive experiments have been carried out, which show that QwT has the following properties:
\begin{itemize}
	\item \textbf{Fast and accurate}. For example, QwT quantizes a ViT network in roughly 2 minutes. During inference, its throughput is almost the same as models quantized by existing quantization methods. QwT is \emph{significantly more accurate} than existing PTQ methods \emph{even without any back-propagation}. On top of that, if higher accuracy is requested, QwT requires \emph{only 1 epoch of training} to approach the accuracy of QAT methods. In contrast, QAT often requires a large number of epochs (\eg, 200 epochs).
	\item \textbf{Simple}. There is \emph{zero (0) hyperparameters} to tune, and the parameters in $S_c$ can be found in \emph{closed-form}.
	\item \textbf{General}. \emph{Exactly the same} QwT method has been successfully applied to various networks and applications, including both CNNs~\cite{resnet} and Transformers (ViT~\cite{vit} and Swin~\cite{swin}), object recognition, detection (with both Mask R-CNN~\cite{mask_rcnn} and DETR~\cite{DETR}) and segmentation, multimodal models (CLIP~\cite{CLIP}), generative models (DiT~\cite{dit}), and large language models (LLaMA~\cite{dubey2024llama3herdmodels}).
	\item \textbf{Practical}. In addition to quantizing to low-bits in simulators, the same QwT method can quantize a model that is able to run directly on GPUs with minimum efforts (\ie, quantization `without tears'): Simply obtain $\theta^{\mathbb Z}$ using TensorRT, then add $S_c$ using QwT. The resulting quantized model is then ready to be deployed on GPUs that support quantized fix-point inference.
\end{itemize}

\section{Related Work}

Network quantization~\cite{quantization_survey} aims to reduce the bit-width of weights and activations, enabling the quantized model to be stored more efficiently and to perform faster inference with suitable hardware support. The fundamental principle of network quantization involves approximating full-precision weights and activations by mapping them to a finite set of discrete values, which are subsequently used in forward model computations (\ie, in inference).

One line of research focuses on quantization-aware training (QAT)~\cite{ReactNet,ewgs,LSQ,Q-ViT}, which integrates quantization into the training process using back-propagation, where the straight-through estimator~\cite{ste} is commonly employed to approximate gradients for non-differentiable rounding functions.

Another line of research concentrates on post-training quantization (PTQ)~\cite{PTQ4ViT,gptq,RepQ-ViT,FQ-ViT}, which converts a fully trained full-precision model into low-bit format using a small set of calibration samples. AdaRound~\cite{adaround} proposed an adaptive weight-rounding mechanism. BRECQ~\cite{brecq} leveraged block reconstruction for quantization, utilizing the Fisher Information Matrix to guide the process. QDrop~\cite{qdrop} randomly dropped the quantization of activations during quantization to achieve flatness of the low-bit model. 

While these methods~\cite{adaround,brecq,qdrop} have proven effective on ResNet~\cite{resnet} backbones, applying them directly to ViT~\cite{vit} often degrades recognition accuracy since the intrinsic structure of the softmax attention is incompatible with these methods. This poses new challenges to design general PTQ methods for the Transformer architecture. 

To address this issue,~\cite{rank_loss} introduced a ranking loss designed to preserve the relative order between quantized and non-quantized attention scores. PTQ4ViT~\cite{PTQ4ViT} proposed twin uniform quantization for shifted activations and a Hessian-guided metric to generate scaling factors. RepQ-ViT~\cite{RepQ-ViT} decoupled quantization and inference, employing distinct quantizers to enable precise quantization while simultaneously ensuring efficient inference. IGQ-ViT~\cite{IGQ-ViT} introduced instance-aware group quantization for ViT to dynamically allocate channels of activation maps to different quantization groups. GPTQ~\cite{gptq} introduced a one-shot weight quantization technique that exploits approximate second-order information.

Different from all of these methods, we propose a new quantization paradigm that introduces a lightweight module to compensate for the information loss caused by quantization. This new paradigm allows our method to be seamlessly integrated with any state-of-the-art quantization methods as a plugin in a completely black-box fashion. Experiments demonstrate that our method is highly compatible with various PTQ approaches, enabling effortless improvements in recognition accuracy within just 2 minutes.

\section{Method}

\subsection{Preliminaries}

We start by outlining key concepts and notations related to network quantization. Given a quantization bit-width $b$, the quantization function $Q(\cdot | b): \mathbb{R} \rightarrow \mathbb{Z}$ maps a floating-point number $x$ (\eg, weight or activation) into its corresponding fixed-point representation $x^{\mathbb{Z}}$ encoded by $b$ bits. Among various quantization approaches, uniform quantization is particularly favored thanks to its simplicity and compatibility with hardware deployment. The uniform quantization procedure is formalized as:
\begin{equation}
    \label{eq:uniform_quant}
    x^{\mathbb{Z}} = \text{clip}\left( \nint{\frac{x}{s}} + z, 0, 2^b - 1 \right) \,,
\end{equation}
in which $s \in \mathbb{R}^+$ represents the quantization scale, and $z \in \mathbb{Z}$ denotes the zero-point offset. These parameters are determined as follows:
\begin{align}
    s &= \frac{\max(x) - \min(x)}{2^b - 1} \,, \label{s} \\
    z &= \text{clip}\left( \nint{-\frac{\min(x)}{s}}, 0, 2^b - 1 \right) \,. \label{z}
\end{align}
In these equations, $\nint{.}$ denotes the rounding function, and the $\text{clip}(\cdot, a, b)$ operation constrains the input value into the range $[a, b]$. The reconstructed quantized output can then be formulated as:
\begin{equation}
    \hat{x} = s \times (x^{\mathbb{Z}} - z) \,.
\end{equation}

Beyond the naive uniform quantizer, a range of more sophisticated quantization techniques~\cite{AdaLog,adaround,PTQ4ViT,RepQ-ViT} have been proposed and extensively studied by the community. In the literature, quantization methods typically quantize both the model weights and activations. 

Quantization significantly reduces the storage requirements by enabling models to be stored in lower bit formats. Additionally, thanks to hardware support for integer-only computations, operations involving the quantized representations $x^{\mathbb{Z}}$, such as matrix multiplications between quantized weights and activations, can be performed with substantial improvements in computational efficiency.

\subsection{Compensation: The Key Insight}

But, obviously there is significant information loss between $x$ and $x^{\mathbb{Z}}$, and it grows very fast when many layers of computation and quantization are stacked together. To recover from the resulting accuracy loss, QAT methods resort to many epochs of training, which leads to the speed-accuracy dilemma and complex, ad hoc quantization methods.

Given a model $M=(S,\theta)$, where $S$ and $\theta$ denote its structure and weights, respectively. Existing quantization techniques transform $M$ into a quantized version $M^{\mathbb Z}=(S^{\mathbb Z},\theta^{\mathbb Z})$, which maintains the same network structure (\ie, $S=S^{\mathbb Z}$) while modifying the original parameters $\theta$ to their quantized counterparts $\theta^{\mathbb Z}$.

Our key insight is to challenge this structural rigidity: the quantized model \emph{does not necessarily need to retain the exactly same structural configuration}, \ie, it is legitimate to allow $S^{\mathbb Z} \neq S$. We argue that some extra modules $S_c$ can be added to the quantized model, such that its structure $S^{\mathbb{Z}}=S \cup S_c$. The extra modules in $S_c$ can \emph{compensate for the information loss caused by quantization}.

More specifically, modern deep nets typically compose of many blocks, \eg, bottleneck blocks in ResNet~\cite{resnet} or Transformer blocks in ViT~\cite{vit}. Let $l_i$ denote the $i$-th block in a model, and let $x_i \in \mathbb R^{d_{in}}$ and $y_i \in \mathbb{R}^{d_{out}}$ be the input and output of this block $l_i$, respectively, such that $y_i = l_i(x_i)$. We argue that we can add a compensation module $c_i$ for this block. Then, $S_c = \bigcup_i c_i$.

For notational simplicity, we omit the subscript $i$ from now on, \ie, we represent a block as $y=l(x)$ and the compensation module is simply denoted as $c$. After quantization, the input activations and weights of the block $l$ are modified into the quantized version $x^{\mathbb{Z}}$ and $l^{\mathbb{Z}}$, respectively. The quantized computation becomes $y^{\mathbb{Z}} = l^{\mathbb{Z}} (x^{\mathbb{Z}})$.

Clearly, there is information loss in all 3 quantization pairs: $l \mapsto l^{\mathbb{Z}}$, $x \mapsto x^{\mathbb{Z}}$ and $y \mapsto y^{\mathbb{Z}}$. What is intriguing is that the information loss is obviously highly non-linear in all 3 pairs. To implement the compensation idea, we have to answer the following questions:
\begin{enumerate}
	\item How to measure the information losses in all 3 pairs that interplay with each other in a complex manner?
	\item How to design the compensation module $c$ that accounts for these highly non-linear information losses?
\end{enumerate}

\subsection{QwT: Quantization without Tears}

We propose a QwT (quantization without tears) method, which answers both questions in the simplest possible form.

First, the information loss is measured by $\|y-y^{\mathbb{Z}}\|^2$. Because $y$ is the output of $l$, it naturally takes care of information losses in $x^{\mathbb{Z}}$ and $l^{\mathbb{Z}}$---when $y^{\mathbb{Z}}=y$, intuitively there is absolutely zero information loss even if $x \neq x^{\mathbb{Z}}$ and $l \neq l^{\mathbb{Z}}$. Note that $\|y-y^{\mathbb{Z}}\|^2$ also accounts for \emph{cumulative} information losses. That is, in the $i$-th block, $c_i$ compensates information losses accumulated in all previous blocks that have not yet been corrected by $c_1, c_2, \dots, c_{i-1}$.

Second, because of this cumulative nature of our choice, in QwT we choose to implement $c$ (index $i$ omitted) as \emph{a simple linear layer}. Although it is impossible to accurately compensate the non-linear information loss in one block via a linear layer, we have many chances to repeatedly apply linear corrections. The entire compensation formed by all extra modules is in fact non-linear because it interacts with the quantized network in every block.

\begin{figure}
	\centering
	\includegraphics[width=0.9\columnwidth]{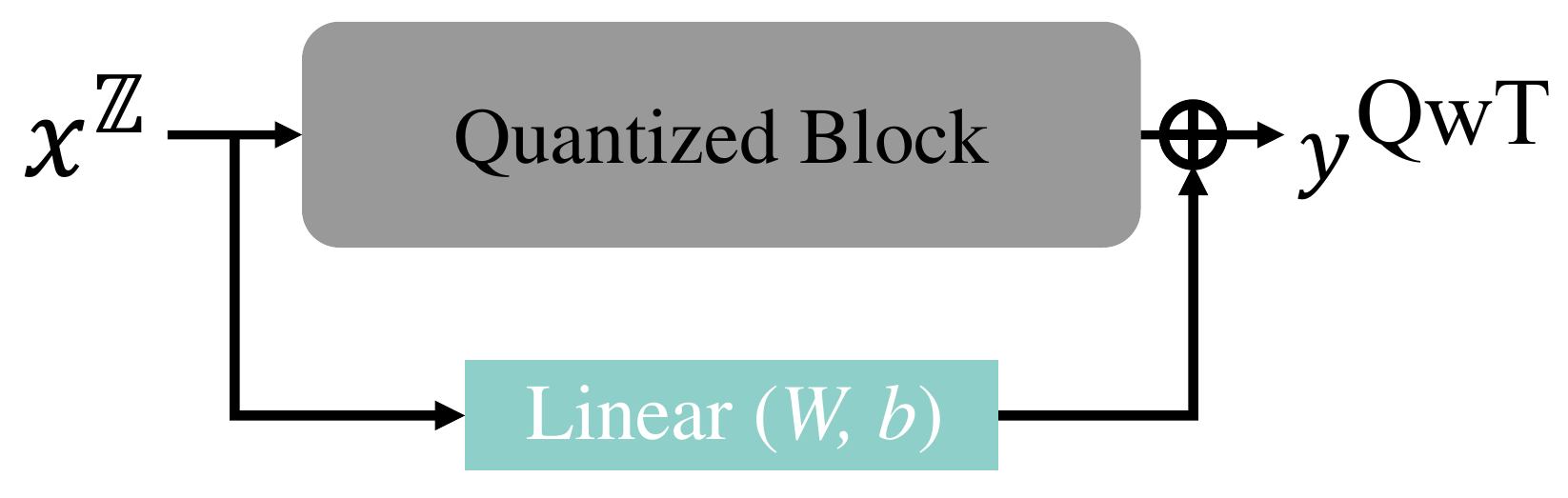}
	\caption{Illustration of QwT in one block. QwT adds a simple linear layer to any model block, compensating for information loss using the block input $x^{\mathbb{Z}}$. This approach is straightforward and compatible with almost all types of backbones~\cite{vit,swin,resnet}.}
	\label{fig:qwt}
\end{figure}

To be concrete, we define $c(x)=Wx+b$ and then 
\begin{equation}
    \label{eq:qwt}
    y^{\text{QwT}}  = l^{\mathbb{Z}} (x^{\mathbb{Z}}) + c(x^{\mathbb{Z}}) \,,
\end{equation}
where $W \in \mathbb{R}^{ d_{out} \times d_{in}}$ and $b \in \mathbb{R}^{d_{out}}$ are the weight matrix and bias vector of the linear layer $c$, respectively. The QwT structure is illustrated in Figure~\ref{fig:qwt}.

These choices are deliberate. They not only make QwT conceptually the simplest, but also ensures a \emph{close-form} solution. To minimize the difference between $y$ and $y^{\mathbb{Z}}$, we select a small set of training examples (512 images) from the training set. Using these samples, QwT collects all inputs for the block $l^{\mathbb{Z}}$ to form a matrix $X^{\mathbb{Z}} \in \mathbb{R}^{d_{in} \times N}$, where $N$ is the total number of features or tokens. We feed $X^{\mathbb{Z}}$ into the quantized block $l^{\mathbb{Z}}$ to obtain $Y^{\mathbb{Z}} \in \mathbb{R}^{d_{out} \times N}$. Next, We feed $X^{\mathbb{Z}}$ into the non-quantized block $l$ to obtain $Y \in \mathbb{R}^{d_{out} \times N}$. Correspondingly, $Y$ contains output of the matching block in the original model. 

Our task is then to \emph{estimate $Y-Y^{\mathbb{Z}}$ using $X^{\mathbb{Z}}$}. This is a classic linear regression problem, and has a closed-form solution:
\begin{equation}
    W^* = (Y-Y^{\mathbb{Z}}) {X^{\mathbb{Z}}}^\top (X^{\mathbb{Z}} {X^{\mathbb{Z}}}^\top)^{-1} \,. \label{eq:W_star}
\end{equation}

In Equation~\ref{eq:W_star}, to simplify the solution, the bias $b$ is absorbed into $W$ and a row vector of $\mathbf{1}^\top$ are concatenated to $X^{\mathbb Z}$. It is worth mentioning that theoretically QwT will \emph{not} make the quantized network getting worse---by setting $W$ and $b$ to all zeros, QwT will not alter the quantized network.

Note that after the QwT module for $l^{\mathbb{Z}}$ is inserted, the compensation in the next block depends on all the previous QwT modules. Consequently, the information loss from $y$ to $y^{\mathbb{Z}}$ is gradually compensated block by block, allowing $c$ to account for the accumulated loss from all preceding blocks that remain uncorrected.

In our experiments, we observed that the coefficient of determination ($R^2$)~\cite{for_r2} for a small subset ($<$5\%) of QwT modules was notably low, adversely hurting recognition accuracy. Consequently, we apply the initialization using Equation~\ref{eq:W_star} only when $R^2>0$; Otherwise, the $W$ and $b$ of the QwT module are set to zero.

Finally, the QwT method has a simple pipeline: first quantize a model $M$ using any PTQ method, then add the compensation module $c_i$ to every block $i$ and set the parameters in $c_i$ using Equation~\ref{eq:W_star}.

A notable advantage of QwT lies in its inherent simplicity. This simplicity ensures that the initialization process of QwT modules is highly efficient, which requires \textit{roughly 2 minutes} in practice to compensate for the information loss during quantization, thereby enhancing recognition accuracy. Experimental results demonstrated that QwT exhibits significant versatility and efficiency across various vision and language tasks.

\section{Experiments}

\begin{table}
\centering
\small
\caption{Quantization results on the ImageNet dataset~\cite{imagenet}. '\#Bits' indicates the bit-width of weights/activations. 'Size' (MB) represents the storage cost of the model on the hard disk. '*' denotes QwT modules and classification head are finetuned for one epoch. '$\dagger$' indicates that the previous state-of-the-art results are directly sourced from the papers~\cite{IGQ-ViT,CL-Calib} due to the unavailability of their official code implementations.}
\begin{tabular}{llccc}
\toprule
Network                       & Method                          & \#Bits       & \phantom{0}Size       & Top-1       \\ \hline
\multirow{9}{*}{DeiT-T}       & Full-precision                   & 32/32        & \phantom{0}22.9        & 72.2        \\ \cdashline{2-5}
                              & IGQ-ViT$^\dagger$~\cite{IGQ-ViT} &  4/4         & -                      & 62.5        \\  
                              & RepQ-ViT~\cite{RepQ-ViT}         &  4/4         & \phantom{00}3.3        & 58.2        \\ 
                              & RepQ-ViT + QwT                   &  4/4         & \phantom{00}4.2        & 61.4        \\
                              & RepQ-ViT + QwT$^*$               &  4/4         & \phantom{00}4.2        & \bf{64.8}   \\ \cdashline{2-5}
                              & IGQ-ViT$^\dagger$~\cite{IGQ-ViT} &  6/6         & -                      & 71.2        \\  
                              & RepQ-ViT~\cite{RepQ-ViT}         &  6/6         & \phantom{00}4.6        & 71.0        \\ 
                              & RepQ-ViT + QwT                   &  6/6         & \phantom{00}5.5        & 71.2        \\
                              & RepQ-ViT + QwT$^*$               &  6/6         & \phantom{00}5.5        & \bf{71.6}   \\
                              \hline
\multirow{9}{*}{Swin-T}       & Full-precision                   & 32/32        & 113.2                 & 81.4        \\ \cdashline{2-5}  
                              & IGQ-ViT$^\dagger$~\cite{IGQ-ViT} &  4/4         & -                     & 77.8        \\ 
                              & RepQ-ViT~\cite{RepQ-ViT}         &  4/4         & \phantom{0}14.9       & 73.0        \\ 
                              & RepQ-ViT + QwT                   &  4/4         & \phantom{0}19.2       & 75.5        \\
                              & RepQ-ViT + QwT$^*$               &  4/4         & \phantom{0}19.2       & \bf{79.3}   \\ \cdashline{2-5}
                              & IGQ-ViT$^\dagger$~\cite{IGQ-ViT} &  6/6         & -                     & 80.9        \\ 
                              & RepQ-ViT~\cite{RepQ-ViT}         &  6/6         & \phantom{0}21.7       & 80.6        \\ 
                              & RepQ-ViT + QwT                   &  6/6         & \phantom{0}26.0       & 80.7        \\
                              & RepQ-ViT + QwT$^*$               &  6/6         & \phantom{0}26.0       & \bf{80.9}   \\
                              \hline
\multirow{9}{*}{ViT-B}        & Full-precision                   & 32/32        & 346.3                 & 84.5        \\ \cdashline{2-5}
                              & IGQ-ViT$^\dagger$~\cite{IGQ-ViT} &  4/4         & -                     & \bf{79.3}   \\  
                              & RepQ-ViT~\cite{RepQ-ViT}         &  4/4         & \phantom{0}44.9       & 68.5        \\ 
                              & RepQ-ViT + QwT                   &  4/4         & \phantom{0}59.1       & 76.3        \\
                              & RepQ-ViT + QwT$^*$               &  4/4         & \phantom{0}59.1       & 78.5        \\ \cdashline{2-5}
                              & IGQ-ViT$^\dagger$~\cite{IGQ-ViT} &  6/6         & -                     & 83.8        \\ 
                              & RepQ-ViT~\cite{RepQ-ViT}         &  6/6         & \phantom{0}66.2       & 83.6        \\ 
                              & RepQ-ViT + QwT                   &  6/6         & \phantom{0}80.4       & 83.9        \\
                              & RepQ-ViT + QwT$^*$               &  6/6         & \phantom{0}80.4       & \bf{84.0}   \\
                              \hline
\multirow{9}{*}{ResNet-50}    & Full-precision                   & 32/32        & 102.2                 & 76.6        \\ \cdashline{2-5}  
                              & CL-Calib$^\dagger$~\cite{CL-Calib} &  4/4       & -                     & \bf{75.4}        \\
                              & Percentile\cite{Percentile}      &  4/4         & \phantom{0}14.0       &  62.3       \\ 
                              & Percentile + QwT                 &  4/4         & \phantom{0}16.0       &  68.5       \\
                              & Percentile + QwT$^*$             &  4/4         & \phantom{0}16.0       &  72.5       \\ \cdashline{2-5}
                              & CL-Calib$^\dagger$~\cite{CL-Calib} &  6/6       & -                     & -           \\  
                              & Percentile\cite{Percentile}      &  6/6         & \phantom{0}19.9       &  76.4       \\ 
                              & Percentile + QwT                 &  6/6         & \phantom{0}21.9       &  76.6       \\
                              & Percentile + QwT$^*$             &  6/6         & \phantom{0}21.9       &  \bf{76.6}           \\
\bottomrule
\end{tabular}
\label{tab:imagenet}
\end{table}

In this section, we begin by evaluating our QwT method on a range of discriminative tasks, including image classification, object detection, instance segmentation, and multimodal recognition. Subsequently, we extend our analysis to generative tasks, such as image generation using diffusion models~\cite{dit} and text generation with large language models~\cite{dubey2024llama3herdmodels}.

\subsection{Experiments on Image Classification}

\begin{table}
\centering
\small
\caption{Results of 8-bit quantization, using tensor-wise Percentile~\cite{Percentile} as the baseline PTQ method. `Latency` (ms) is measured on a single RTX 3090 GPU with a batch size of 64, utilizing Nvidia's TensorRT~\cite{TensorRT} toolkit for deployment.}
\begin{tabular}{llccc}
\toprule
Network                       & Method               & \phantom{0}Size       & Latency          & Top-1    \\ \hline
\multirow{3}{*}{DeiT-T}       & Full-precision       & \phantom{0}22.9       & 11.6             & 72.2     \\ 
                              & Percentile~\cite{Percentile}           & \phantom{00}5.9       & \phantom{0}2.8   & 71.2     \\
                              & Percentile + QwT     & \phantom{00}6.8       & \phantom{0}3.2   & 71.5     \\
\hline                              
\multirow{3}{*}{Swin-T}       & Full-precision       & 113.2                 & 34.5             & 81.4     \\
                              & Percentile~\cite{Percentile}           & \phantom{0}28.6       & \phantom{0}9.5   & 80.8     \\
                              & Percentile + QwT     & \phantom{0}32.9       & 10.9             & 81.0     \\
\hline                              
\multirow{3}{*}{Swin-S}       & Full-precision       & 198.4 	             & 61.0             & 83.2     \\
                              & Percentile~\cite{Percentile}           & \phantom{0}50.1       & 16.0             & 82.1     \\
                              & Percentile + QwT     & \phantom{0}58.0       & 17.9             & 83.0     \\
\hline
\multirow{3}{*}{ViT-S}        & Full-precision       & \phantom{0}88.2       & 28.3             & 81.4     \\
                              & Percentile~\cite{Percentile}           & \phantom{0}22.5       & \phantom{0}5.8   & 79.2     \\
                              & Percentile + QwT     & \phantom{0}26.0       & \phantom{0}6.6   & 80.1     \\
\hline
\multirow{3}{*}{ViT-B}        & Full-precision       & 346.3                 & 85.3             & 84.5     \\
                              & Percentile~\cite{Percentile}           & \phantom{0}87.4       & 15.5             & 75.8     \\
                              & Percentile + QwT     & 101.6                 & 17.5             & 82.8     \\
\bottomrule
\end{tabular}
\label{tab:inference_latency}
\end{table}

\textbf{Settings.} We evaluated our method on image classification tasks using the ImageNet dataset~\cite{imagenet}, leveraging various backbone architectures including ViT~\cite{vit}, DeiT~\cite{deit}, Swin~\cite{swin}, and ResNet~\cite{resnet}. We randomly sampled 512 images from the training set to initialize the parameters of the QwT modules using Equation~\ref{eq:W_star}. In all networks, the affine transformation matrix $W$ in QwT is implemented in FP16 format to reduce model size. In ResNet, $W$ is further simplified as a group-wise convolution using a kernel size of 1 and 64 channels per group, achieving additional efficiency in storage and computation. Note that a group-wise convolution is still a linear operator, which can be perfectly encoded by the pair $(W,b)$. Other details were consistent with prior work~\cite{RepQ-ViT}. Please refer to the appendix for more information.

\textbf{Results on different backbones.} Table~\ref{tab:imagenet} summarizes the quantization results when applying QwT across different backbone architectures. Specifically, we selected RepQ-ViT~\cite{RepQ-ViT} and Percentile~\cite{Percentile} as the baseline methods for the Transformer family~\cite{vit,deit,swin} and ResNet~\cite{resnet}, respectively. The results show that incorporating the QwT module consistently boosts the recognition accuracy, leading to an average increase of approximately 2.6\%, and even up to 5\% for 4-bit quantization, highlighting that QwT is particularly effective for low-bit scenarios. Additionally, after the QwT modules are integrated, the accuracy in 6-bit quantization cases aligns closely with prior state-of-the-art approaches~\cite{IGQ-ViT}.

The potential of our QwT method can be further unlocked through finetuning. By jointly optimizing the QwT modules and the classification head for only one (1) additional epoch (results marked with~$^*$), more gains in accuracy are achieved, enabling our method to surpass previous state-of-the-art results in nearly all cases. 

These results are even closer to those produced by QAT methods, which typically require extensive training (e.g., 200 epochs). In contrast, our QwT$^*$ achieves similar performance with only one epoch of finetuning. Our approach not only substantially improves training efficiency but also keeps the backbone parameters unchanged, making it more suitable for hardware deployment.

In Table~\ref{tab:inference_latency}, we additionally report the inference latency of different models directly deployed on a GPU. Compared to full-precision models, naive quantized models achieve an average reduction of 77\% in inference latency and 75\% in model size. When QwT modules are incorporated, these reductions slightly decrease to 74\% and 71\%, respectively, with an overhead of only 3\%. This minimal additional cost is offset by a average 1.9 percentage points improvement in recognition accuracy, demonstrating the strong practicality of the QwT method.

\textbf{Results across various PTQ methods.} We extended our experiments to evaluate the versatility of QwT by applying it to various PTQ methods. As shown in Table~\ref{tab:different_ptq_method}, we integrated QwT into PTQ4ViT~\cite{PTQ4ViT}, RepQ-ViT~\cite{RepQ-ViT}, and Percentile~\cite{Percentile}, using ViT-B as the backbone. 

We observe that QwT consistently enhances top-1 accuracy across all baseline PTQ methods. Notably, in 4-bit scenarios, PTQ4ViT demonstrates an improvement of approximately 40\%, while RepQ-ViT shows an 8\% increase. Compared to modern PTQ methods~\cite{RepQ-ViT, IGQ-ViT, PTQ4ViT}, which often involve complex and tedious procedures, our method demonstrates high simplicity and, most importantly, is compatible with all these approaches, too. The significant improvement in accuracy narrows the performance gap between different PTQ methods, and offers new insights into the design of new paradigms for network quantization.

\begin{table}
\centering
\small
\caption{Quantization results among different PTQ methods on the ImageNet dataset~\cite{imagenet} using ViT-B~\cite{vit} as the backbone.}
\begin{tabular}{lccc}
\toprule
Method                          & \#Bits       & Size                  & Top-1       \\ \hline
Full-precision                  & 32/32        & 346.3                 & 84.5        \\ \hline
PTQ4ViT~\cite{PTQ4ViT}          & 4/4          & \phantom{0}44.9       & 30.7        \\
PTQ4ViT+QwT                     & 4/4          & \phantom{0}59.1       & \textbf{70.0}        \\ \cdashline{1-4}
RepQ-ViT~\cite{RepQ-ViT}        & 4/4          & \phantom{0}44.9       & 68.5        \\ 
RepQ-ViT + QwT                  & 4/4          & \phantom{0}59.1       & \textbf{76.3}        \\ \hline
Percentile~\cite{Percentile}    & 6/6          & \phantom{0}66.2       & 56.7         \\
Percentile+QwT                  & 6/6          & \phantom{0}80.4       & \textbf{79.8}       \\  \cdashline{1-4}
PTQ4ViT~\cite{PTQ4ViT}          & 6/6          & \phantom{0}66.2       & 81.7        \\
PTQ4ViT+QwT                     & 6/6          & \phantom{0}80.4       & \textbf{83.2}        \\ \cdashline{1-4}
RepQ-ViT~\cite{RepQ-ViT}        & 6/6          & \phantom{0}66.2       & 83.6        \\ 
RepQ-ViT + QwT                  & 6/6          & \phantom{0}80.4       & \textbf{83.9}        \\ \hline
Percentile~\cite{Percentile}    & 8/8          & \phantom{0}87.4       & 75.8        \\
Percentile+QwT                  & 8/8          & 101.6      & \textbf{82.8}       \\ 
\bottomrule
\end{tabular}
\label{tab:different_ptq_method}
\end{table}

\begin{table}[t]
\centering
\small
\caption{Quantization results of applying QwT finetuning schema on QAT methods.}
\begin{tabular}{llcc}
\toprule
Network                    & Method      & \#Bits      & Top-1       \\ \hline
\multirow{4}{*}{DeiT-S}    & Q-ViT~\cite{Q-ViT}        &  2/2         & 72.1   \\ 
                           & Q-ViT + QwT$^*$           &  2/2         & \textbf{72.5}   \\ \cdashline{2-4}
                           & Q-ViT~\cite{Q-ViT}        &  3/3         & 79.0   \\ 
                           & Q-ViT + QwT$^*$           &  3/3         & \textbf{79.1}   \\
\bottomrule
\end{tabular}
\label{tab:qat_finetune}
\end{table}

\textbf{Extension to QAT methods.} We further investigated the potential of adapting QwT to QAT methods. Specifically, we applied QwT modules to QAT models after completing QAT training to assess whether QwT can further enhance recognition accuracy. 

We preliminarily found that for QAT models, the initialization process described by Equation~\ref{eq:W_star} is no longer effective. Applying it directly to QAT models significantly degrades accuracy. We attribute this to the fact that, unlike full-precision models, the optimization state of a QAT-trained model is sufficiently converged, resulting in almost no information loss from $y$ to $y^{\mathbb Z}$. In fact, $y^{\mathbb Z}$ may even outperform $y$, as QAT models sometimes surpass their full-precision counterparts in evaluation accuracy. 

To integrate QwT into QAT methods, we therefore initialize $W$ and $b$ to zero as a compromise. We then explore whether fine-tuning QwT can still improve recognition accuracy. For this study, we use Q-ViT~\cite{Q-ViT}, a representative QAT method for ViT backbones, as the baseline. The results in Table~\ref{tab:qat_finetune} demonstrate that, even without using the initialization from Equation~\ref{eq:W_star}, fine-tuning the QwT modules consistently enhances QAT models, confirming the generalizability of our approach.

\subsection{Experiments on Object Detection \& Instance Segmentation}

\textbf{Settings.} We evaluated our method on object detection and instance segmentation tasks using the COCO 2017~\cite{COCO} dataset. ResNet50~\cite{resnet} with DETR~\cite{DETR}, Swin-S~\cite{swin} with Mask R-CNN~\cite{mask_rcnn}, and Swin-S/B~\cite{swin} with Cascade Mask R-CNN~\cite{cascade_r_cnn} were used as detectors. The evaluation metric was Average Precision (AP). Similar to image classification, we randomly selected 512 images from the training set to initialize the QwT weights and biases. For ResNet, the QwT was implemented using group-wise convolution with a kernel size of 1 and 64 channels per group to balance model size and AP. For DETR, we used MinMax as the PTQ baseline, a classic method that quantizes the model based on the range between the minimum and maximum values of weights or activations. For the other detectors, RepQ-ViT~\cite{RepQ-ViT} was chosen as the baseline PTQ method.

\textbf{Main results.} Table~\ref{tab:coco} presents the results of applying QwT to object detection and instance segmentation tasks. We observe that QwT consistently enhances both $\text{AP}^{\text{box}}$ and $\text{AP}^{\text{mask}}$ across \emph{all} cases \emph{without finetuning}, achieving an average improvement of 0.4\% with individual gains ranging from 0.1\% to 0.7\%. The consistent improvement underscores the robustness of our method for both object detection and instance segmentation tasks. Notably, in certain 6-bit scenarios, such as on Cascade Mask R-CNN, QwT even achieves AP comparable to full-precision models.

Additionally, a clear trend emerges where the AP gains introduced by QwT increases along with model size. For instance, in $\text{AP}^{\text{box}}$, the average improvement achieved by QwT rises from 0.3\% in ResNet-50+DETR to 0.5\% in Swin-B+Cascade Mask R-CNN, indicating the method's enhanced effectiveness in larger models.

Compared to full-precision models, baseline PTQ methods yield an average storage reduction of approximately 80\%. The introduction of QwT modules slightly reduces this savings to around 78\% (-2\%), which demonstrates that QwT enhances AP metrics with negligible overhead.

\begin{table}
\centering
\small
\setlength{\tabcolsep}{2.4pt}
\caption{Quantization results on the COCO dataset~\cite{imagenet}. We use box average precision ($\text{AP}^{\text{box}}$) and mask average precision ($\text{AP}^{\text{mask}}$) to assess object detection and instance segmentation accuracy, respectively.}
\setlength{\tabcolsep}{0.5mm}
\begin{tabular}{llcccc}
\toprule
Network                       & Method           & \#Bits       & Size       & $\text{AP}^{\text{box}}$    & $\text{AP}^{\text{mask}}$  \\ \hline
\multirow{5}{*}{\begin{tabular}[l]{@{}l@{}}\quad ResNet-50\\ \ \ \ \ \ + DETR \end{tabular}}       & Full-precision                   & 32/32        & 164.5       & 42.0  & -  \\  \cdashline{2-6}
                              & MinMax                        &  6/6         & \phantom{0}47.4        & 39.5     & -      \\ 
                              & MinMax + QwT                  &  6/6         & \phantom{0}49.4        & \textbf{40.0}     & -      \\ \cdashline{2-6}
                              & MinMax                        &  8/8         & \phantom{0}56.4        & 41.6     & -      \\ 
                              & MinMax + QwT                  &  8/8         & \phantom{0}58.4        & \textbf{41.7}     & -      \\
                              \hline
\multirow{5}{*}{\begin{tabular}[l]{@{}l@{}}\quad \quad Swin-S\\ + Mask R-CNN \end{tabular}}       & Full-precision                   & 32/32        & 276.5  & 48.5  & 43.3 \\ \cdashline{2-6}
                              & RepQ-ViT~\cite{RepQ-ViT}   &  4/4         & \phantom{0}36.1        & 42.6     & 40.0      \\ 
                              & RepQ-ViT + QwT             &  4/4         & \phantom{0}44.0        & \textbf{43.1}     & \textbf{40.4}      \\ \cdashline{2-6}
                              & RepQ-ViT~\cite{RepQ-ViT}   &  6/6         & \phantom{0}53.3        & 47.6     & 42.9      \\ 
                              & RepQ-ViT + QwT             &  6/6         & \phantom{0}61.2        & \textbf{48.0}     & \textbf{43.1}      \\
                              \hline
\multirow{5}{*}{\begin{tabular}[l]{@{}l@{}}\quad \ \ Swin-S\\ \ \ \ + Cascade \\ \ \ Mask R-CNN \end{tabular}}       & Full-precision                   & 32/32        & 427.8  & 51.9  & 45.0 \\ \cdashline{2-6}
                              & RepQ-ViT~\cite{RepQ-ViT}   &  4/4         & \phantom{0}56.9        & 49.3     & 43.1      \\ 
                              & RepQ-ViT + QwT             &  4/4         & \phantom{0}64.8        & \textbf{49.9}     & \textbf{43.4}      \\ \cdashline{2-6}
                              & RepQ-ViT~\cite{RepQ-ViT}   &  6/6         & \phantom{0}83.4        & 51.4     & 44.6      \\ 
                              & RepQ-ViT + QwT             &  6/6         & \phantom{0}91.3        & \textbf{51.7}     &  \textbf{44.8}      \\
                              \hline
\multirow{5}{*}{\begin{tabular}[l]{@{}l@{}}\quad \ \ Swin-B\\ \ \ \ + Cascade \\ \ \ Mask R-CNN \end{tabular}}       & Full-precision                   & 32/32        & 579.9  & 51.9  & 45.0 \\ \cdashline{2-6}
                              & RepQ-ViT~\cite{RepQ-ViT}   &  4/4         & \phantom{0}76.1         & 49.3     & 43.1      \\ 
                              & RepQ-ViT + QwT             &  4/4         & \phantom{0}90.1         & \textbf{50.0}     & \textbf{43.7}      \\ \cdashline{2-6}
                              & RepQ-ViT~\cite{RepQ-ViT}   &  6/6         & 112.1        & 51.5     & 44.8      \\ 
                              & RepQ-ViT + QwT             &  6/6         & 126.1        & \textbf{51.8}     & \textbf{45.0}      \\
\bottomrule
\end{tabular}
\label{tab:coco}
\end{table}

\subsection{Experiments on Multimodal Recognition}

\textbf{Settings.} We conducted experiments using OpenAI's CLIP model~\cite{CLIP}. Known for its exceptional zero-shot performance on the ImageNet~\cite{imagenet} classification task, CLIP serves as an ideal benchmark for assessing the effectiveness on multimodal recognition tasks. We selected the variant of CLIP that includes a ViT-B/32~\cite{vit} as the visual encoder and a 12-block Transformer~\cite{Transformer} as the text encoder. Since, to the best of our knowledge, no publicly available PTQ implementation exists for CLIP, we developed a baseline using RepQ-ViT~\cite{RepQ-ViT}. We randomly selected 512 image-text pairs from the training data, both for PTQ model calibration and QwT initialization. Thanks to the simplicity and efficiency of our method, it achieved significant improvements under 30 seconds, as detailed in Table~\ref{tab:clip}.

\textbf{Main results.} We conducted experiments with two quantization strategies: quantizing 1) only the visual encoder and 2) both visual and text encoders. As shown in Table~\ref{tab:clip}, baseline PTQ methods showed significant drop in top-1 accuracy compared to their full-precision counterparts, struggling to effectively represent a low-bit CLIP model. The reduction in performance is especially obvious when both the visual and text encoders are quantized.

In contrast, our QwT method enhanced top-1 accuracy across all cases, significantly bridging the accuracy gap between low-bit and full-precision models. Specifically, in vision-only quantization, QwT increased top-1 accuracy by an average of 0.6\%, with only a modest 4\% increase in model size compared to baseline PTQ methods.

When both the visual and text encoders are quantized, baseline PTQ methods exhibited an average accuracy drop of 29.2\%. In contrast, QwT provided a significant accuracy improvement, with an average increase of 14.8\%. These findings highlight QwT's effectiveness in preserving high accuracy while substantially reducing model size for multimodal recognition tasks.

\begin{table}
\centering
\small
\setlength{\tabcolsep}{4pt}
\caption{Quantization results of CLIP for zero-shot classification tasks on ImageNet. The 'Quant Setup' column differentiates between two strategies: \textit{quantizing only the Visual Encoder} and \textit{quantizing both the Visual and Text Encoders concurrently}.}
\begin{tabular}{llccc}
\toprule
Quant Setup                       & Method                           & \#Bits       & Size (MB)       & Top-1  \\ \hline
\multirow{5}{*}{Vision}       & Full-precision                  & 32/32             & 607.2       & 63.4  \\ \cdashline{2-5}
                              & RepQ-ViT~\cite{RepQ-ViT}        &  6/6              & 323.5       & 59.2  \\ 
                              & RepQ-ViT + QwT                  &  6/6              & 336.8       & \textbf{60.3}  \\ \cdashline{2-5}
                              & RepQ-ViT~\cite{RepQ-ViT}        &  8/8              & 345.3       & 62.9  \\ 
                              & RepQ-ViT + QwT                  &  8/8              & 359.5       & \textbf{63.0}  \\
                              \hline
\multirow{5}{*}{\begin{tabular}[l]{@{}c@{}}Vision  \\ \& Text \end{tabular}}       & Full-precision                   & 32/32        & 607.2  & 63.4  \\ \cdashline{2-5}
                              & RepQ-ViT~\cite{RepQ-ViT}        &  6/6             & 200.8        & 29.8      \\ 
                              & RepQ-ViT + QwT                  &  6/6             & 221.3        & \textbf{43.5}      \\ \cdashline{2-5}
                              & RepQ-ViT~\cite{RepQ-ViT}        &  8/8             & 232.1        & 38.7      \\ 
                              & RepQ-ViT + QwT                  &  8/8             & 252.6        & \textbf{54.6}      \\

\bottomrule
\end{tabular}
\label{tab:clip}
\end{table}

\subsection{Experiments on Image Generation}

\begin{figure*}[t!]
    \centering
    \includegraphics[width=0.9\textwidth]{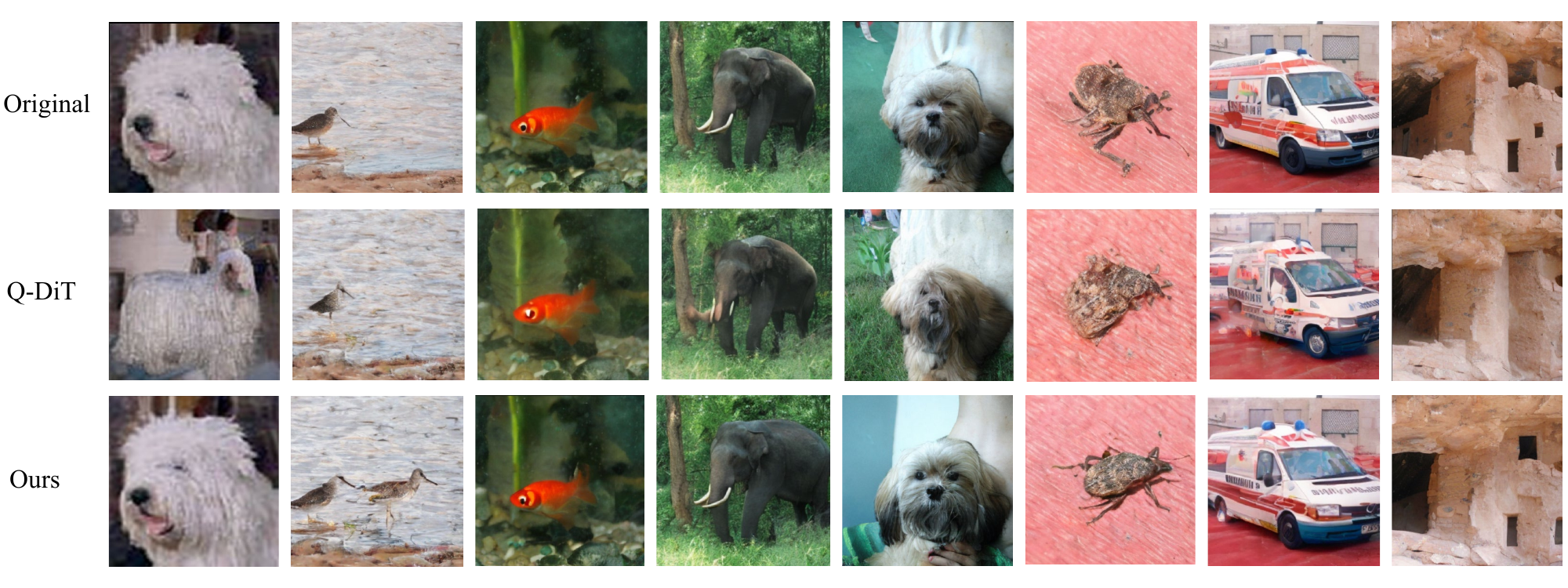}
    \caption{Qualitative visualization results of quantizing DiT-XL/2.}
    \label{fig:dit}
\end{figure*}

QwT has also demonstrated efficacy in generative models, notably enhancing the performance of quantized diffusion models. Unlike classifiers or detectors, which require a single forward pass, diffusion models involve multiple forward passes to generate the final images, presenting a unique prototype. Under these circumstances, QwT has proven itself highly effective, underscoring its general applicability and robustness.

\textbf{Settings}. For our experiments, we selected the influential DiT~\cite{dit} (Diffusion Transformer) architecture, following the experimental setup of Q-DiT~\cite{Q-DiT}. Specifically, we employed pretrained DiT-XL/2 models at a resolution of 256×256. For rapid and precise sampling, we utilized the DDIM sampler with 50 sampling steps and applied classifier-free guidance (cfg) of 1.5, abbreviated as DiT-XL/2 (steps = 50, cfg = 1.5). Our experiments included two quantization configurations: W8A8 and W4A8. Additional results involving various model sizes, steps, and cfg values are available in the appendix.

We applied QwT directly to the quantized diffusion model using Q-DiT. A key consideration is that the model performs $T$ forward passes per inference, with notable variation in the activation distribution and range across steps. A key assumption is that quantization error is primarily dependent on the input $x$, with minimal influence from elements like the time step or class condition. Accordingly, we set $t=0$ to initialize the compensation module. The results are presented in Table~\ref{tab:dit}.

\textbf{Main results.} Our method was compared with three representative quantization techniques: RepQ-ViT, GPTQ and Q-DiT designed for diffusion models. For both W8A8 and W4A8 settings, QwT significantly enhanced the performance of the quantized models, yielding improvements of 0.10 and 0.69 in FID, which illustrates the efficacy of QwT with minimal increase in model size.

We visualize the images generated by our model alongside those from compared models in Figure~\ref{fig:dit}. The three rows represent the original images, quantized images with Q-DiT, and quantized images with QwT, respectively. All models are based on DiT-XL/2 (steps = 50, cfg = 1.5). To enable a fair comparison, we ensure that the initial Gaussian noise and the noise added during inference are identical across all methods. The images produced by our method show a closer visual resemblance to the original model, which aligns with the quantitative results.

\begin{table}
    \centering
    \caption{Quantitative results of quantizing DiT-XL/2. $\downarrow$ ($\uparrow$) means smaller (larger) is better.}
	\setlength{\tabcolsep}{3pt}
    \begin{tabular}{lcccc}
        \toprule
        Method & \#Bits & Size (MB) & \phantom{0}FID ($\downarrow$) & IS ($\uparrow$) \\
        \midrule
        Full-precision & 16/16  & 1349 & \phantom{00}5.32 & 236.17 \\
        \midrule
        RepQ-ViT & 8/8 & \phantom{0}677 & \phantom{00}5.46 & 234.74 \\
        GPTQ & 8/8 & \phantom{0}690 & \phantom{00}5.90 & 218.90 \\
        Q-DiT& 8/8 & \phantom{0}683 & \phantom{00}5.45 & 236.52 \\
        Q-DiT + QwT &8/8 & \phantom{0}707 & \phantom{00}\textbf{5.35} & \textbf{236.91} \\
        \midrule
        RepQ-ViT & 4/8 & \phantom{0}339 & 319.68 & \phantom{00}2.20 \\
        GPTQ & 4/8 & \phantom{0}351 & \phantom{00}9.94 & 166.35 \\
        Q-DiT & 4/8 & \phantom{0}347 & \phantom{00}6.75 & 208.38 \\
        Q-DiT + QwT & 4/8 & \phantom{0}361 & \phantom{00}\textbf{6.06} & \textbf{215.70} \\
        \bottomrule
    \end{tabular}
    \label{tab:dit}
\end{table}

\subsection{Experiments on Large Language Models}

\textbf{Settings.} We evaluated our framework on the LLaMA3-8B~\cite{dubey2024llama3herdmodels} model. For PTQ methods, we adopted GPTQ~\cite{gptq} with INT4 weight quantization. Our approach is also compatible with other PTQ methods such as AWQ~\cite{lin2023awq} and SPQR~\cite{dettmers2024spqr}. We conducted a group-wise asymmetric quantization with a group size of 128 and apply activation reordering. In particular, GPTQ take 128 samples from the C4 dataset as calibration sets, and each sample is 2048 tokens long. We use the same calibration set when performing QwT after GPTQ algorithm.

\textbf{Evaluation metrics.} Following the settings of GPTQ, we evaluated the perplexity on the WikiText2~\cite{stephen2017pointer} and C4~\cite{raffel2020exploring} datasets. We further assessed the zero-shot commonsense question answering (QA) ability on eight tasks covering SIQA~\cite{sap2019social}, HellaSwag~\cite{zellers2019hellaswag}, PIQA~\cite{Bisk2020piqa}, WinoGrande~\cite{sakaguchi2021winogrande}, ARC~\cite{clark2018think}, BoolQ~\cite{clark2019boolq}, and OpenBookQA~\cite{mihaylov2018can}. We also evaluated both the zero-shot and five-shot performance of the LLMs on Massively Multitask Language Understanding (MMLU) benchmark~\cite{hendrycks2021mmlu}. It consists of 57 language tasks including humanities, STEM, social science, \etc. We adopted lm-eval-harness~\cite{gao2021framework} to produce the accuracy results.     

\textbf{Results}. Table~\ref{tab:llm_ptq_method} summarizes the perplexity in WikiText2, C4 and the average accuracy in eight common sense reasoning datasets. More results are shown in the appendix.  Note that we abbreviate WikiText2 to W2. As the results show, our optimized models will not overfit the calibration dataset and consistently outperform the original PTQ models. These results reveal the effectiveness of our QwT.

\begin{table}[t]
    \centering
    \small
    \setlength{\tabcolsep}{3pt}
    \caption{Quantization results among WikiText2, C4 and eight zero-shot commonsense QA datasets using LLaMA3-8B as the backbone. $\downarrow$ ($\uparrow$) means smaller (larger) is better.}
    \begin{tabular}{lccccc}
    \toprule
    Method & \#Bits & Size (GB) &  W2 ($\downarrow$) & C4 ($\downarrow$) & QA. Avg ($\uparrow$)     \\ \hline
    Full-precision     & 16  & 16.06 & 6.24 & 8.96 & 66.10 \\ \hline
    GPTQ    & 4   & \phantom{0}5.73 & 6.65 & 9.44 &  64.90  \\
    GPTQ + QwT     & 4   & \phantom{0}6.80 & \textbf{6.63} & \textbf{9.38}  & \textbf{65.18}\\ 
    \bottomrule
    \end{tabular}
    \label{tab:llm_ptq_method}
    \end{table}

\section{Conclusions}
In this paper, we proposed Quantization without Tears (QwT), a novel approach that incorporates a lightweight structure into quantized models to compensate for the information loss during network quantization. The QwT modules, implemented as a tiny set of linear layers and seamlessly integrated into backbone blocks, achieved accuracy, simplicity, and generality simultaneously. Notably, QwT provides a closed-form solution to complete the compensation process in under 2 minutes and enables effortless integration with existing quantization techniques. Extensive experiments demonstrated QwT's exceptional effectiveness and versatility across a wide range of tasks, models, and quantization methods, advancing a streamlined and flexible paradigm for network quantization.

\section*{Contributions}

J.W. designed the compensation insight and the QwT framework mathematically. M.F. made them into algorithms and codes that work well in practice, and carried out the main empirical validations. H.Y., J.S. and J.Z. carried out experiments and validations on LLM, AIGC and multimodal tasks, respectively. K.Z. engaged in discussions. All authors contributed to paper writing.

\section*{Correction Note (July 2025)}

In the originally published camera-ready version of \href{https://openaccess.thecvf.com/content/CVPR2025/papers/Fu_Quantization_without_Tears_CVPR_2025_paper.pdf}{this paper}, an issue in the implementation of the Percentile PTQ baseline led to incorrect configuration labels in Table~\ref{tab:imagenet} for the ResNet-50 backbone. The results reported under ``4/4'' and ``6/6'' bit-width settings were actually obtained using ``4/8'' and ``6/8'' configurations, respectively. We have now corrected Table~\ref{tab:imagenet} to display the right ``4/4'' and ``6/6'' results. Furthermore, in Table~\ref{tab:resnet}, we have extended our experiments to include all weight and activation bit-width combinations (``4/4'', ``4/8'', ``6/6'', and ``6/8'') for completeness. These updates do not affect any of the paper's conclusions. We apologize for any confusion caused.

{
    \small
    \bibliographystyle{ieeenat_fullname}
    \bibliography{qwerty_ref}
}

\clearpage
\appendix

\section{Full Implementation Details}

In this section, we present full implementation details of the different types of tasks in our experiments.

\textbf{Image Classification.} We selected RepQ-ViT~\cite{RepQ-ViT}, PTQ4ViT~\cite{PTQ4ViT}, and Percentile~\cite{Percentile} as the primary baseline PTQ methods to integrate with our QwT modules. Following~\cite{RepQ-ViT}, we randomly sampled 32 images from the ImageNet~\cite{imagenet} dataset as the calibration set to initialize the quantized weights for these baseline methods. Additionally, a separate set of 512 randomly selected images from the ImageNet training set was used to initialize the parameters of the QwT modules (excluding PTQ weights). For all networks, the affine transformation matrix \( W \) in QwT is implemented in FP16 format to reduce model size. In ResNet, \( W \) is further simplified as a group-wise convolution with a kernel size of 1 and 64 channels per group. 

When finetuning the QwT modules along with the classification head for an additional epoch, we utilized AdamW~\cite{adamw} as the optimizer. The batch size was set to 32 per GPU (using a total of 4 GPUs), and weight decay was set to 0. The learning rate was configured to 1e-7 for ViT~\cite{vit}, 5e-6 for DeiT~\cite{deit} and Swin~\cite{swin}, 1e-5 for ResNet~\cite{resnet}. 

In addition to the original classification loss, during 1-epoch finetuning we applied a simple distillation loss to minimize the squared L2-distance between the full-precision and quantized models—calculated on the output features before the classification head (\textit{cls token} for ViT and DeiT, \textit{global average feature} for Swin and ResNet), yielding the finetuning objective as $L_{cls} + L_{dis}$ (\ie, without the combination weight hyperparameter.) 

This distillation strategy is the feature mimicking method~\cite{wang2021distilling}, which only utilizes the penultimate features and argues that features or activations from intermediate layers are not necessary or even harmful. It is also worth noting since only the penultimate features are required, feature mimicking is unsupervised.

\textbf{Object Detection \& Instance Segmentation.} Following~\cite{RepQ-ViT}, we randomly sampled a single image from the COCO dataset~\cite{COCO} to initialize the quantized weights for baseline PTQ methods. All other details are consistent with the image classification case.

\textbf{Image Generation.} Consistent with the experimental setup of Q-DiT~\cite{Q-DiT}, we selected the DiT architecture and employed pretrained DiT-XL/2 models at a resolution of $256 \times 256$. Our experiments were extended to a broader range of settings, including varying the number of sampling steps (50 and 100) and classifier-free guidance (CFG) scales (0 and 1.5). These results are presented in the next section.

\section{More Experimental Results}

In this section, we provide more comprehensive quantization results across a range of backbones~\cite{vit,deit,swin,resnet} on the ImageNet dataset~\cite{imagenet}, as summarized in Tables~\ref{tab:vit_deit}, \ref{tab:swin}, and \ref{tab:resnet}.

We also show the full results for large language models in Table~\ref{tab:appendix_detailed_llm_ptq_method}. The main text only reports the overall average accuracy on eight zero-shot commonsense QA datasets. Table~\ref{tab:appendix_detailed_llm_ptq_method} lists the accuracy for each dataset separately. We also include the results on the MMLU benchmarks, tested in both zero-shot and five-shot modes.

\begin{figure*}
    \centering
    \includegraphics[width=0.95\linewidth]{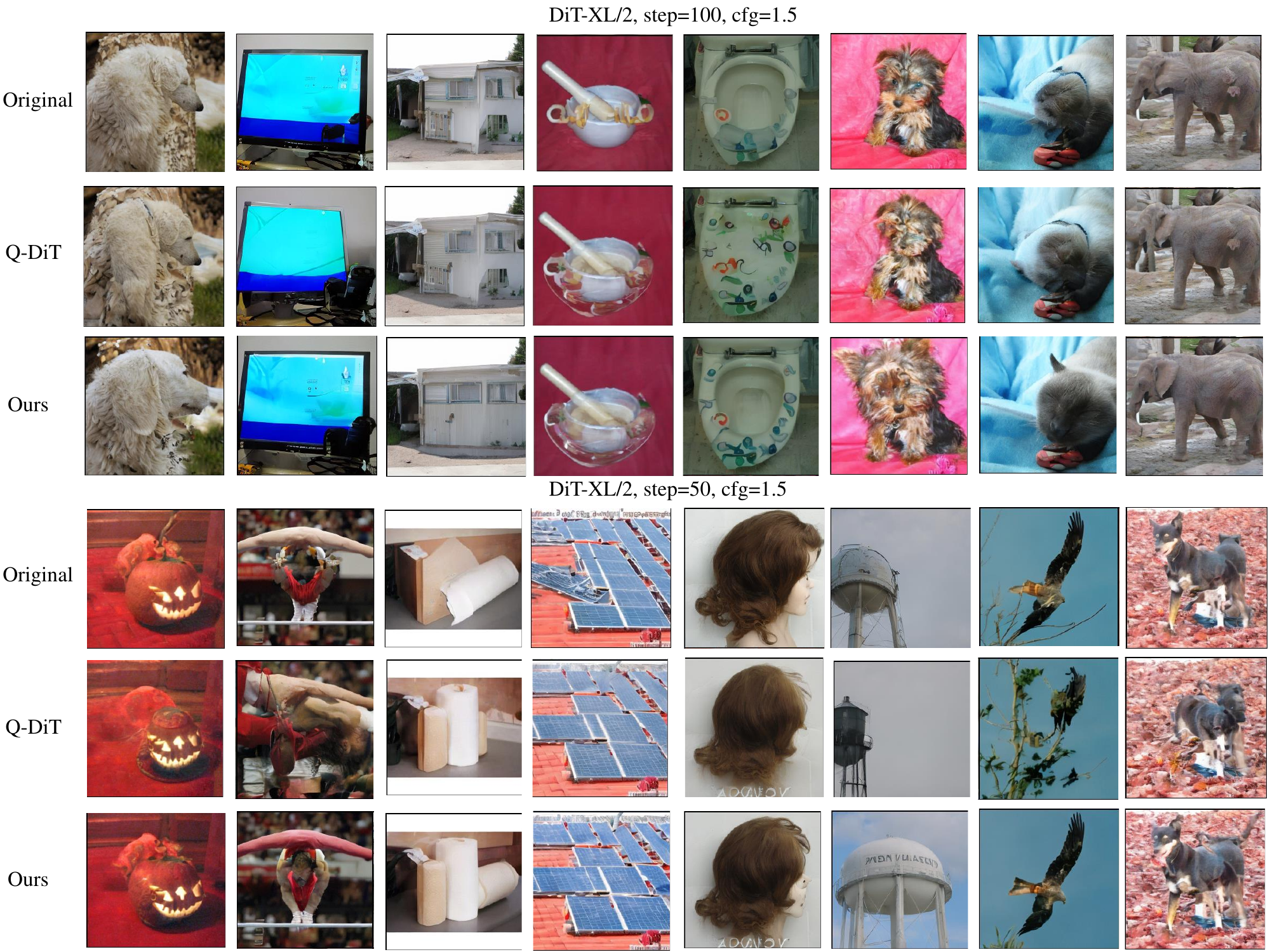}
    \caption{Qualitative visualization results of quantizing DiT-XL/2 on different settings.}
    \label{fig:visual_appendix}
\end{figure*}

\begin{figure*}
    \centering
    \includegraphics[width=0.9\textwidth]{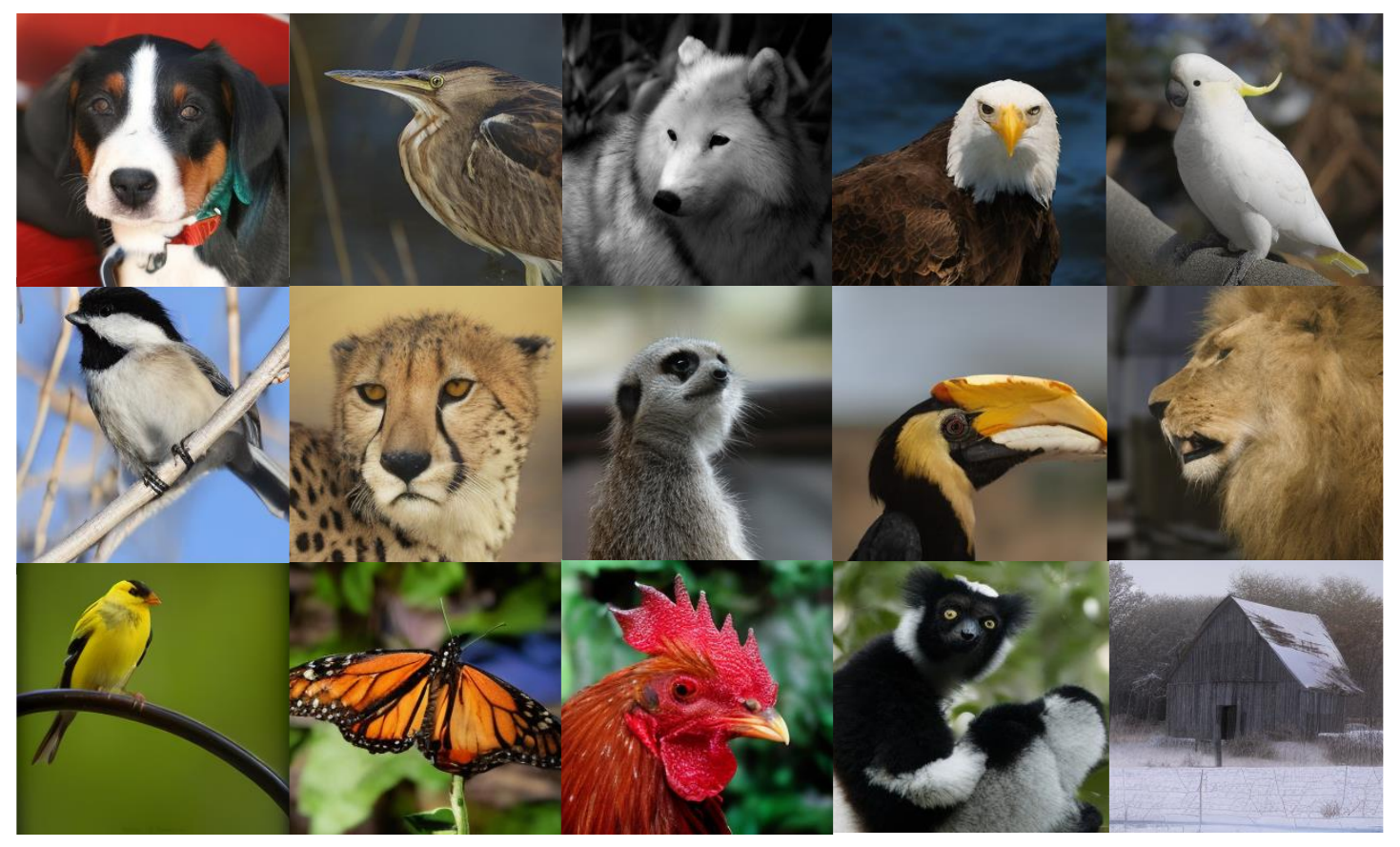}
    \caption{More qualitative visualization results of our method on quantized DiT-XL/2.}
    \label{fig:dit_appendix}
\end{figure*}

The full results of image generation are summarized in~\Cref{tab:results}. As shown in the table, our method consistently enhances the performance of the generative model across all tested configurations. To provide a more intuitive understanding, we visualize the generated images under each setting in~\Cref{fig:visual_appendix}. Similar to the main paper, we ensured that the noise during the generation process remains consistent across all models. The visualizations further confirm that our method reliably improves the quality of the generated images. As a further illustration, we provide several representative images generated by our method in \Cref{fig:dit_appendix}.

\begin{table}
\centering
\small
\caption{Full results on ViT~\cite{vit} and DeiT~\cite{deit} backbones.}
\begin{tabular}{llccc}
\toprule
Network                       & Method                          & \#Bits       & \phantom{0}Size       & Top-1       \\ \hline
\multirow{9}{*}{DeiT-T}       & Full-precision                   & 32/32        & \phantom{0}22.9      & 72.2        \\ \cdashline{2-5}
                              & IGQ-ViT$^\dagger$~\cite{IGQ-ViT} &  4/4         & -                    & 62.5        \\  
                              & RepQ-ViT~\cite{RepQ-ViT}         &  4/4         & \phantom{00}3.3      & 58.2        \\ 
                              & RepQ-ViT + QwT                   &  4/4         & \phantom{00}4.2      & 61.4        \\
                              & RepQ-ViT + QwT$^*$               &  4/4         & \phantom{00}4.2      & \bf{64.8}   \\ \cdashline{2-5}
                              & IGQ-ViT$^\dagger$~\cite{IGQ-ViT} &  6/6         & -                    & 71.2        \\  
                              & RepQ-ViT~\cite{RepQ-ViT}         &  6/6         & \phantom{00}4.6      & 71.0        \\ 
                              & RepQ-ViT + QwT                   &  6/6         & \phantom{00}5.5      & 71.2        \\
                              & RepQ-ViT + QwT$^*$               &  6/6         & \phantom{00}5.5      & \bf{71.6}   \\
                              \hline
\multirow{9}{*}{DeiT-S}       & Full-precision                   & 32/32        & \phantom{0}88.2      & 79.9        \\ \cdashline{2-5}
                              & IGQ-ViT$^\dagger$~\cite{IGQ-ViT} &  4/4         & -                    & 74.7        \\  
                              & RepQ-ViT~\cite{RepQ-ViT}         &  4/4         & \phantom{0}11.9      & 69.0        \\ 
                              & RepQ-ViT + QwT                   &  4/4         & \phantom{0}15.4      & 71.5        \\
                              & RepQ-ViT + QwT$^*$               &  4/4         & \phantom{0}15.4      & \bf{75.2}   \\ \cdashline{2-5}
                              & IGQ-ViT$^\dagger$~\cite{IGQ-ViT} &  6/6         & -                    & 79.3        \\  
                              & RepQ-ViT~\cite{RepQ-ViT}         &  6/6         & \phantom{0}17.2      & 78.9        \\ 
                              & RepQ-ViT + QwT                   &  6/6         & \phantom{0}20.7      & 79.1        \\
                              & RepQ-ViT + QwT$^*$               &  6/6         & \phantom{0}20.7      & \textbf{79.3}   \\
                              \hline
\multirow{9}{*}{ViT-S}        & Full-precision                  & 32/32        & 88.2       & 81.4        \\ \cdashline{2-5}
                              & IGQ-ViT$^\dagger$~\cite{IGQ-ViT}&  4/4         & -          & \bf{73.6}   \\  
                              & RepQ-ViT~\cite{RepQ-ViT}        &  4/4         & 11.9       & 65.8        \\ 
                              & RepQ-ViT + QwT                 &  4/4         & 15.4       & 70.8        \\
                              & RepQ-ViT + QwT$^*$             &  4/4         & 15.4       & 72.9        \\ \cdashline{2-5}
                              & IGQ-ViT$^\dagger$~\cite{IGQ-ViT}          &  6/6         & -          & 80.8        \\  
                              & RepQ-ViT~\cite{RepQ-ViT}        &  6/6         & 17.2       & 80.5        \\ 
                              & RepQ-ViT + QwT                 &  6/6         & 20.7       & 80.7        \\
                              & RepQ-ViT + QwT$^*$             &  6/6         & 20.7       & \bf{80.8}   \\
                              \hline
\multirow{9}{*}{ViT-B}        & Full-precision                   & 32/32        & 346.3                 & 84.5        \\ \cdashline{2-5}
                              & IGQ-ViT$^\dagger$~\cite{IGQ-ViT} &  4/4         & -                     & \bf{79.3}   \\  
                              & RepQ-ViT~\cite{RepQ-ViT}         &  4/4         & \phantom{0}44.9       & 68.5        \\ 
                              & RepQ-ViT + QwT                   &  4/4         & \phantom{0}59.1       & 76.3        \\
                              & RepQ-ViT + QwT$^*$               &  4/4         & \phantom{0}59.1       & 78.5        \\ \cdashline{2-5}
                              & IGQ-ViT$^\dagger$~\cite{IGQ-ViT} &  6/6         & -                     & 83.8        \\ 
                              & RepQ-ViT~\cite{RepQ-ViT}         &  6/6         & \phantom{0}66.2       & 83.6        \\ 
                              & RepQ-ViT + QwT                   &  6/6         & \phantom{0}80.4       & 83.9        \\
                              & RepQ-ViT + QwT$^*$               &  6/6         & \phantom{0}80.4       & \bf{84.0}   \\

\bottomrule
\end{tabular}
\label{tab:vit_deit}
\end{table}

\begin{table}
\centering
\small
\caption{Full results on the Swin~\cite{swin} backbone.}
\begin{tabular}{llccc}
\toprule
Network                       & Method                          & \#Bits       & \phantom{0}Size       & Top-1       \\ \hline
\multirow{9}{*}{Swin-T}       & Full-precision                   & 32/32        & 113.2                 & 81.4        \\ \cdashline{2-5}  
                              & IGQ-ViT$^\dagger$~\cite{IGQ-ViT} &  4/4         & -                     & 77.8        \\ 
                              & RepQ-ViT~\cite{RepQ-ViT}         &  4/4         & \phantom{0}14.9       & 73.0        \\ 
                              & RepQ-ViT + QwT                   &  4/4         & \phantom{0}19.2       & 75.5        \\
                              & RepQ-ViT + QwT$^*$               &  4/4         & \phantom{0}19.2       & \bf{79.3}   \\ \cdashline{2-5}
                              & IGQ-ViT$^\dagger$~\cite{IGQ-ViT} &  6/6         & -                     & 80.9        \\ 
                              & RepQ-ViT~\cite{RepQ-ViT}         &  6/6         & \phantom{0}21.7       & 80.6        \\ 
                              & RepQ-ViT + QwT                   &  6/6         & \phantom{0}26.0       & 80.7        \\
                              & RepQ-ViT + QwT$^*$               &  6/6         & \phantom{0}26.0       & \bf{80.9}   \\
                              \hline
\multirow{9}{*}{Swin-S}       & Full-precision                   & 32/32        & 198.4                 & 83.2        \\ \cdashline{2-5}  
                              & IGQ-ViT$^\dagger$~\cite{IGQ-ViT} &  4/4         & -                     & 81.0        \\ 
                              & RepQ-ViT~\cite{RepQ-ViT}         &  4/4         & \phantom{0}25.8       & 80.2        \\ 
                              & RepQ-ViT + QwT                   &  4/4         & \phantom{0}33.7       & 80.4        \\
                              & RepQ-ViT + QwT$^*$               &  4/4         & \phantom{0}33.7       & \bf{81.9}   \\ \cdashline{2-5}
                              & IGQ-ViT$^\dagger$~\cite{IGQ-ViT} &  6/6         & -                     & 82.9        \\ 
                              & RepQ-ViT~\cite{RepQ-ViT}         &  6/6         & \phantom{0}38.0       & 82.8        \\ 
                              & RepQ-ViT + QwT                   &  6/6         & \phantom{0}45.9       & 82.9        \\
                              & RepQ-ViT + QwT$^*$               &  6/6         & \phantom{0}45.9       & \bf{82.9}   \\
\bottomrule
\end{tabular}
\label{tab:swin}
\end{table}

\begin{table}
\centering
\small
\caption{Full results on the ResNet~\cite{resnet} backbone. `\#Bits' indicates the bit-width of weights/activations. `$\dagger$' indicates that the previous state-of-the-art results are directly sourced from the papers~\cite{CL-Calib} due to the unavailability of their official code implementations.}
\begin{tabular}{llccc}
\toprule
Network                       & Method                          & \#Bits       & \phantom{0}Size       & Top-1       \\ \hline
\multirow{9}{*}{ResNet-18}    & Full-precision                   & 32/32        & \phantom{0}46.8                 & 71.0        \\ \cdashline{2-5}  
                              & CL-Calib$^\dagger$~\cite{CL-Calib} &  4/4       & -                     & \bf{69.4}        \\
                              & Percentile\cite{Percentile}      &  4/4         & \phantom{00}6.1       & 47.1        \\ 
                              & Percentile + QwT                 &  4/4         & \phantom{00}6.4       & 62.3        \\
                              & Percentile + QwT$^*$             &  4/4         & \phantom{00}6.4       & 65.5        \\ \cdashline{2-5}
                              & Percentile\cite{Percentile}      &  4/8         & \phantom{00}6.1       & 58.3        \\ 
                              & Percentile + QwT                 &  4/8         & \phantom{00}6.4       & 68.9        \\
                              & Percentile + QwT$^*$             &  4/8         & \phantom{00}6.4       & \bf{69.4}   \\ \cdashline{2-5}
                              & Percentile\cite{Percentile}      &  6/6         & \phantom{00}8.9       & 70.4         \\ 
                              & Percentile + QwT                 &  6/6         & \phantom{00}9.2       & 70.7        \\
                              & Percentile + QwT$^*$             &  6/6         & \phantom{00}9.2       & \bf{71.0}        \\ \cdashline{2-5}
                              & Percentile\cite{Percentile}      &  6/8         & \phantom{00}8.9       & 70.7        \\ 
                              & Percentile + QwT                 &  6/8         & \phantom{00}9.2       & 71.0        \\
                              & Percentile + QwT$^*$             &  6/8         & \phantom{00}9.2       & \bf{71.1}   \\
                              \hline
\multirow{9}{*}{ResNet-50}    & Full-precision                   & 32/32        & 102.2                 & 76.6        \\ \cdashline{2-5}  
                              & CL-Calib$^\dagger$~\cite{CL-Calib} &  4/4       & -                     & \bf{75.4}   \\
                              & Percentile\cite{Percentile}      &  4/4         & \phantom{0}14.0       & 62.3        \\ 
                              & Percentile + QwT                 &  4/4         & \phantom{0}16.0       & 68.5        \\
                              & Percentile + QwT$^*$             &  4/4         & \phantom{0}16.0       & 72.5            \\ \cdashline{2-5}
                              & Percentile\cite{Percentile}      &  4/8         & \phantom{0}14.0       & 68.4        \\ 
                              & Percentile + QwT                 &  4/8         & \phantom{0}16.0       & 74.5        \\
                              & Percentile + QwT$^*$             &  4/8         & \phantom{0}16.0       & \bf{75.8}   \\ \cdashline{2-5}
                              & Percentile\cite{Percentile}      &  6/6         & \phantom{0}19.9       & 76.4        \\ 
                              & Percentile + QwT                 &  6/6         & \phantom{0}21.9       & 76.6        \\
                              & Percentile + QwT$^*$             &  6/6         & \phantom{0}21.9       & \bf{76.6}   \\ \cdashline{2-5}
                              & Percentile\cite{Percentile}      &  6/8         & \phantom{0}19.9       & 76.0        \\ 
                              & Percentile + QwT                 &  6/8         & \phantom{0}21.9       & 76.8        \\
                              & Percentile + QwT$^*$             &  6/8         & \phantom{0}21.9       & \bf{76.8}   \\
                              \hline
\multirow{9}{*}{ResNet-101}    & Full-precision                   & 32/32        & 178.2                 & 77.3       \\ \cdashline{2-5}  
                              & Percentile\cite{Percentile}      &  4/4         & \phantom{0}23.7       &  67.5       \\ 
                              & Percentile + QwT                 &  4/4         & \phantom{0}28.0       &  71.1       \\
                              & Percentile + QwT$^*$             &  4/4         & \phantom{0}28.0       &  \bf{74.5}       \\ \cdashline{2-5}
                              & Percentile\cite{Percentile}      &  4/8         & \phantom{0}23.7       & 74.7        \\ 
                              & Percentile + QwT                 &  4/8         & \phantom{0}28.0       & 76.4        \\
                              & Percentile + QwT$^*$             &  4/8         & \phantom{0}28.0       & \bf{76.7}   \\ \cdashline{2-5}
                              & Percentile\cite{Percentile}      &  6/6         & \phantom{0}34.3       & 76.8        \\ 
                              & Percentile + QwT                 &  6/6         & \phantom{0}38.6       & 76.9        \\
                              & Percentile + QwT$^*$             &  6/6         & \phantom{0}38.6       & \bf{76.9}        \\ \cdashline{2-5}
                              & Percentile\cite{Percentile}      &  6/8         & \phantom{0}34.3       & 77.1        \\ 
                              & Percentile + QwT                 &  6/8         & \phantom{0}38.6       & 77.2        \\
                              & Percentile + QwT$^*$             &  6/8         & \phantom{0}38.6       & \bf{77.2}   \\
\bottomrule
\end{tabular}
\label{tab:resnet}
\end{table}

\begin{table*}
    \centering
    \caption{Detailed quantization results among the MMLU dataset and eight zero-shot commonsense QA datasets using LLaMA3-8B as the backbone.}
    \setlength{\tabcolsep}{0.8mm}
    \small
    \begin{tabular}{lc|cc|ccccccccc}
      \hline
      Method &  \#Bits &  MMLU (0-shot) & MMLU (5-shot) & BoolQ & PIQA & SIQA & HLSW & WG &ARC-e & ARC-c & OBQA & QA. Avg \\ 
      \hline
      Full-precision & 16 & 63.39 & 65.30 & 82.17 & 81.18 & 32.91 & 78.93 & 73.95 & 81.14 & 53.50 & 45.00 & 66.10\\  \hline
      GPTQ & 4 & 61.40 & 63.94 & 81.25 & 81.39 & 32.91 & 78.28 & 72.77 & 78.03 & 50.60 & 44.00 & 64.90\\  
      GPTQ + QwT & 4 & \textbf{61.57} & \textbf{64.25} & 81.22 & 81.45 & 32.91 & 77.77 & 73.40 & 79.21 & 50.68 & 44.80 & \textbf{65.18}\\          \hline
    \end{tabular}
    \label{tab:appendix_detailed_llm_ptq_method}
  \end{table*}

\begin{table*}
    \centering
    \caption{Quantitative results of quantizing DiT-XL/2 on ImageNet $256\times 256$.}
    \label{tab:results}
    \resizebox{\textwidth}{!}{
    \begin{tabular}{lllcccccc}
    \toprule
    \textbf{Model}           & \textbf{Bit-width (W/A)} & \textbf{Method} & \textbf{Size (MB)} & \textbf{FID (↓)} & \textbf{sFID (↓)} & \textbf{IS (↑)} & \textbf{Precision (↑)} &  \textbf{Recall (↑)} \\ 
    \midrule
    \multirow{6}{*}{DiT-XL/2 (steps = 100)} 
    & 16/16 & FP         & 1349  & 12.40 & 19.11 & 116.68 & 0.6605 & - \\
    \cmidrule{2-9}
    &\multirow{5}{*}{4/8}& PTQ4DM     & 339   & 252.31 & 82.44 & 2.74   & 0.0125 & - \\
    &       & RepQ-ViT   & 339   & 315.85 & 139.99 & 2.11   & 0.0067 & - \\
    &       & GPTQ       & 351   & 25.48 & 25.57 & 73.46  & 0.5392 & - \\
    &       & Q-DiT      & 347   & 15.76 & 19.84 & 98.78  & \textbf{0.6395} & - \\ 
    &       & Q-DiT + QwT& 361 & \textbf{15.35} & \textbf{19.63} & \textbf{104.04} & 0.6373 & \textbf{0.7478} \\ 
    \midrule
    \multirow{6}{*}{DiT-XL/2 (steps = 100, cfg = 1.5)} 
    & 16/16 & FP         & 1349  & 5.31  & 17.61 & 245.85 & 0.8077 & - \\
    \cmidrule{2-9}
    & \multirow{5}{*}{4/8}   & PTQ4DM     & 339   & 255.06 & 84.63 & 2.76   & 0.0110 & - \\
    &       & RepQ-ViT   & 339   & 311.31 & 138.58 & 2.18   & 0.0072 & - \\
    &       & GPTQ       & 351   & 7.66  & 20.76 & 193.76 & 0.7261 & - \\
    &       & Q-DiT & 347   & 6.40  & 18.60 & 211.72 & 0.7609 & - \\ 
    &       & Q-DiT + QwT& 361 & \textbf{5.86} & \textbf{18.29} & \textbf{221.66} & \textbf{0.7678} & \textbf{0.6915} \\
    \midrule
    \multirow{6}{*}{DiT-XL/2 (steps = 50)} 
    & 16/16 & FP         & 1349  & 13.47 & 19.31 & 114.71 & 0.6601 & -\\
    \cmidrule{2-9}
    & \multirow{5}{*}{4/8}   & PTQ4DM     & 339   & 256.15 & 83.45 & 2.73   & 0.0150 & - \\
    &       & RepQ-ViT   & 339   & 324.25 & 142.98 & 2.12   & 0.0062 & - \\
    &       & GPTQ       & 351   & 26.31 & 25.54 & 69.73  & 0.5388 & - \\
    &       & Q-DiT & 347   & 17.42 & 19.95 & 97.52  & 0.6219  & -\\ 
    &       & Q-DiT + QwT& 361 & \textbf{17.02} & \textbf{19.57} & \textbf{99.62} & \textbf{0.6302} & \textbf{0.7582} 
    \\
    \bottomrule
    \end{tabular}
    }
    \end{table*}

\end{document}